\documentclass[11pt]{article}
\pdfoutput=1

\usepackage{arxiv}
\usepackage[utf8]{inputenc}
\usepackage[T1]{fontenc}
\usepackage{newtxtext,newtxmath}
\usepackage[hidelinks]{hyperref}
\usepackage{url}
\usepackage{booktabs}
\usepackage{multirow}
\usepackage{graphicx}
\usepackage{amsmath}
\usepackage{microtype}
\usepackage[numbers,sort&compress]{natbib}
\usepackage{xcolor}
\usepackage{enumitem}
\usepackage{caption}
\setlist{nosep,leftmargin=1.4em}

\newcommand{\repourl}{\url{https://github.com/piyushdhoka/depth-pruned-slm-recovery}}


\title{What Breaks and What Comes Back:\\Capability Damage and Recovery in Depth-Pruned Small Language Models}
\renewcommand{\shorttitle}{Capability Damage and Recovery in Depth-Pruned Small Language Models}

\author{%
  Piyush Dhoka \And Vedant Pandhare \And Varun Inamdar \And Aadarsh Pathre \And Pavitha Nooji \AND
  \normalfont Department of Artificial Intelligence, Vishwakarma University, Pune, India}

\date{October 2026}

\begin{document}
\maketitle

\begin{abstract}
Depth pruning removes whole Transformer blocks to make a language model smaller and faster, but the damage is not
spread evenly across what the model can do, and a short fine-tuning run, called healing, decides what comes back.
We measure this for small models used as agents. Four instruction-tuned models of 1.2--1.7B parameters
(Llama-3.2-1B, Qwen2.5-1.5B, SmolLM2-1.7B and OLMo-2-1B) are pruned and healed, and six capabilities are tracked:
structured output, function calling, instruction following, mathematical reasoning, safety refusal and knowledge.
Three results hold across the families. First, function calling and math collapse in every model (retention
0.00--0.07 of the unpruned score), and the order in which capabilities break agrees closely across the four
(Kendall's $W = 0.92$; 0.62 after correcting for chance floors). Second, which parameters are trained matters as much as which data: fully tuning the last
three blocks instead of adding low-rank adapters to all blocks costs up to 0.18 in capped mean retention on three
families, and one 300k-token pilot per scope detects this. Third, we test a planner that allocates a 1M-token
healing budget from 21 short pilot heals. At a fixed budget it raises the weakest capability on two families,
although only one of these gains survives a test that also resamples training seeds. Once its 3.48M pilot tokens are
counted, uniform healing with the same total number of training tokens matches or beats every planned variant on
every family and retains math significantly better on three of them in an item-level test (on one when seeds are
also resampled). Healing longer is a strong and simple baseline that planned recovery must
beat. Code, per-item predictions and every result are public.
\end{abstract}


\section{Introduction}

Small instruction-tuned language models with one to two billion parameters run on phones, laptops and inexpensive
servers, and they are increasingly used as \emph{agents}: they must return valid JSON, call tools with the right
arguments, follow formatting instructions and refuse unsafe requests. Every parameter still costs memory, bandwidth
and energy, so compression matters. Depth pruning, which deletes entire Transformer blocks, is attractive because the
pruned model needs no special kernels and every removed block takes its share of parameters, key--value cache and
per-token compute with it~\citep{men2025shortgpt,gromov2025unreasonable,kim2024shortened,song2024sleb}.

Depth pruning does not damage capabilities evenly. Removing only two of the sixteen blocks of Llama-3.2-1B-Instruct
leaves function calling at 0\% of its original score and structured output at 6\%, while instruction following
keeps 43\%. The
standard remedy is \emph{healing}: a short fine-tuning run on a modest amount of
data~\citep{gromov2025unreasonable,kim2024shortened,thangarasa2025selfdata}. Healing has a budget, usually counted in
training tokens, and that budget has to be split across data that target different capabilities. Most work heals on
one generic corpus; recent methods adapt the data mixture~\citep{xia2024sheared,li2025datamixing} or allocate
recovery data to whatever has degraded most~\citep{he2026paser}.

This paper asks four practical questions about sub-2B instruction-tuned models:
\begin{enumerate}
  \item Which capabilities break when the model is depth-pruned, and is the pattern the same across model families?
  \item Does it matter \emph{which parameters} are trained during healing, and not only which data?
  \item Does planning the healing mixture from short pilot heals beat simple allocation rules at a fixed budget?
  \item Does that advantage survive once the training tokens spent on the pilots are counted?
\end{enumerate}

\begin{figure}[t]
  \centering
  \includegraphics[width=\linewidth]{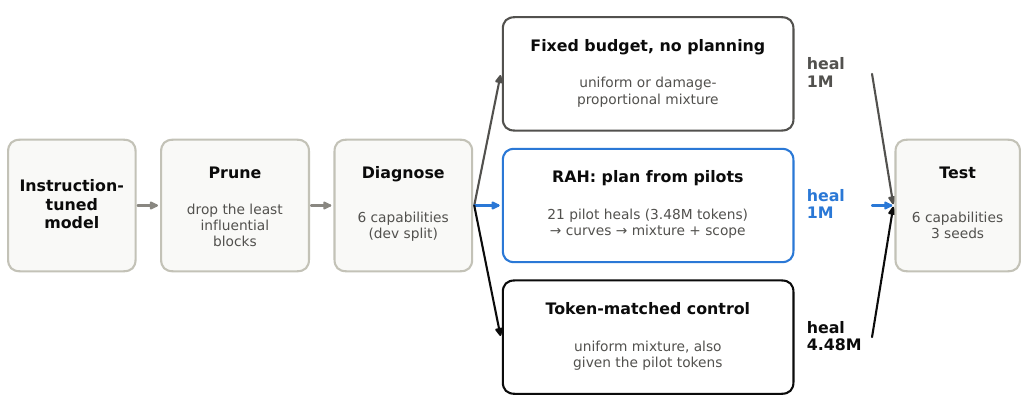}
  \caption{Study design. Within a family, every lane uses the same pruned model, data pools, teacher targets, seeds
  and test items, and one training recipe per adaptation scope. RAH spends 3.48M training tokens on 21 pilot heals before its 1M-token heal. We compare it with
  the top lane at the same 1M-token healing budget, and with the bottom lane, which receives the same 4.48M total
  training tokens.}
  \label{fig:design}
\end{figure}

To answer the last two questions we use recoverability-aware healing (RAH), a planner in the style of recent
data-mixing optimizers~\citep{li2025datamixing,zhang2026domainpilot}: it runs 21 short pilot heals, fits a recovery
curve per capability and solves for the mixture and the adaptation scope. RAH is the object under test, not a method
we claim is better. The key control is \emph{token-matched}: a uniform heal that receives the planner's pilot tokens
as extra training. Figure~\ref{fig:design} summarizes the design.

\paragraph{Findings.}
\begin{itemize}
  \item \textbf{What breaks.} Function calling and mathematical reasoning collapse in all four families. The order in
  which the six capabilities break agrees closely across families ($W = 0.92$); after correcting for chance floors
  the agreement is only moderate ($W = 0.62$), and the order beyond function calling and math depends on the family.
  \item \textbf{How you heal matters.} Fully tuning the last three blocks matches low-rank adaptation of all blocks on
  one family and loses 0.05--0.18 in capped mean retention on the other three. One 300k-token pilot per scope picks
  the better option in every family.
  \item \textbf{Planning at a fixed budget.} With 1M healing tokens, the planned mixture never lowers capped mean
  retention significantly, and it raises the weakest capability (math) by up to 0.17 on Llama and 0.14 on OLMo
  (RAH-capped). Only the OLMo gain survives the seed-aware test, and on Llama the gain is confounded with a change
  of adaptation scope.
  \item \textbf{Planning once its cost is counted.} A uniform heal that also receives the 3.48M pilot tokens is never
  significantly worse than any planned variant. It retains math significantly better on Qwen, SmolLM2 and OLMo in
  the item-level test, and on SmolLM2 also in the seed-aware test. The pilot-fitted curves cannot see how slowly math
  recovers, so they misjudge where extra tokens pay off.
\end{itemize}

\paragraph{Contributions.} (i) A six-capability diagnosis of depth pruning for sub-2B instruction-tuned models,
aimed at agent use. (ii) Measurements of capability-specific recovery and of the effect of adaptation scope under
identical data. (iii) An accounting of planning cost: planned allocations are compared both at a fixed healing
budget and against a uniform heal that receives the same total number of training tokens, with seed-aware
statistics. (iv) A full release of code, data recipes, per-item predictions and results at \repourl. Most Llama
and Qwen runs were done on 8\,GB laptop GPUs, so the study can be reproduced without a cluster.

\section{Related Work}

\paragraph{Depth pruning and recovery.} ShortGPT~\citep{men2025shortgpt} scores each block by its Block Influence
(one minus the cosine similarity between its input and output) and removes the lowest-scoring blocks.
\citet{gromov2025unreasonable} remove a contiguous span chosen by angular distance and heal with QLoRA on generic
text. Other methods refine the choice of blocks~\citep{song2024sleb,lu2026reassessing}, replace removed spans with
light modules~\citep{chen2025streamlining,shopkhoev2025replaceme}, or recover without training by patching hidden
states~\citep{chen2025linearpatch,feng2025restoring}. Retraining comparisons exist for LoRA versus continued
pre-training~\citep{kim2024shortened} and for partial-layer fine-tuning~\citep{lu2026reassessing}; scaling laws for
post-pruning training have been proposed~\citep{chen2025p2law}, and retraining can matter more than the pruning
criterion~\citep{wagner2025freelunch}. Most of this work reports perplexity or multiple-choice accuracy for models
of 7B parameters or more.

\paragraph{Allocating recovery data.} PASER~\citep{he2026paser} clusters recovery instructions, scores how much
each cluster has degraded, allocates the data budget in proportion and selects samples within each cluster. Our
damage-proportional baseline adopts its allocation principle with task-score damage, not the full PASER pipeline.
Data-mixture optimization is well established for pre-training~\citep{xie2023doremi,xia2024sheared} and has recently
been applied to supervised fine-tuning by fitting small mixture runs~\citep{li2025datamixing,zhang2026domainpilot}.
RAH carries that pilot-and-fit idea over to healing.

\paragraph{What breaks under compression.} Compression harms math and generative reasoning more than
classification~\citep{shrestha2026limits,jaiswal2024compressing}, tool use can break before intent
recognition~\citep{zhang2026smarthomes}, and agentic abilities degrade in ways perplexity does not
show~\citep{dong2025acbench}. Compression also shifts safety and refusal behavior~\citep{xu2024beyond,hong2024decoding},
and width pruning of Llama-3.2 can leave instruction following intact while losing
knowledge~\citep{martra2025fragile}. These studies diagnose damage; we also measure recovery and what it costs.

\section{Study Design}

\subsection{Models and pruning}

We use four open instruction-tuned families (Table~\ref{tab:models}): Llama-3.2-1B-Instruct~\citep{grattafiori2024llama3}
(through an ungated public mirror of the official weights), Qwen2.5-1.5B-Instruct~\citep{qwen2024qwen25},
SmolLM2-1.7B-Instruct~\citep{benallal2025smollm2} and OLMo-2-1B-Instruct~\citep{olmo2025olmo2}. Each model is pruned
by Block Influence~\citep{men2025shortgpt}, computed on 126 teacher-generated healing examples (21 per pool) cut into
256-token chunks. The first token position is excluded because it acts as an attention sink~\citep{xiao2024attentionsinks},
and block~0 is never removed. Kept blocks are re-indexed so that key--value caching stays correct.

The number of removed blocks is a per-family design choice, not a common fraction. A useful diagnosis needs
capabilities that are damaged but still measurable: removing four of Llama's sixteen blocks left five of six
capabilities at their chance floor, so Llama and OLMo lose two blocks, Qwen loses seven of 28 and SmolLM2 four of 24.
The results therefore compare the \emph{order} in which capabilities break within each family, not damage levels
across families. A lighter cut is evaluated for diagnosis only.

\begin{table}[t]
  \centering
  \footnotesize
  \setlength{\tabcolsep}{2.2pt}
  \caption{Model families, parameters (billions), Block-Influence cuts (original block indices) and the GPUs the
  runs used.}
  \label{tab:models}
  \begin{tabular}{llrrll}
    \toprule
    Family & Checkpoint & Params & Blocks & Removed & GPU \\
    \midrule
    Llama-3.2-1B & \texttt{unsloth/Llama-3.2-1B-Instruct} & 1.24 & 16 & 12, 13 (12.5\%) & RTX 4060 (laptop); L4 \\
Qwen2.5-1.5B & \texttt{Qwen/Qwen2.5-1.5B-Instruct} & 1.54 & 28 & 13--16, 24--26 (25.0\%) & RTX 4060 (laptop); L4 \\
SmolLM2-1.7B & \texttt{HuggingFaceTB/SmolLM2-1.7B-Instruct} & 1.71 & 24 & 6, 20--22 (16.7\%) & A100 \\
OLMo-2-1B & \texttt{allenai/OLMo-2-0425-1B-Instruct} & 1.48 & 16 & 8, 13 (12.5\%) & A100 \\
    \bottomrule
  \end{tabular}
\end{table}

\subsection{Capability suite}

Table~\ref{tab:suite} lists the six capabilities. Items are fixed, seeded subsets, so every checkpoint of a family
sees identical inputs. The development and test items are disjoint: development items are used only by the pilots
and by the damage estimate that the allocation rules read, and test items only for the reported numbers. Generation
is greedy with per-capability length limits and each model's own chat template. Structured output asks for a JSON
object valid under a JSONSchemaBench~\citep{geng2025jsonschemabench} schema. Function calling uses BFCL v3 simple and
multiple-function items~\citep{patil2025bfcl}, scored by abstract-syntax-tree match of the function name and
arguments. Instruction following uses IFEval's strict prompt-level checkers~\citep{zhou2023ifeval} through
lm-evaluation-harness~\citep{gao2023lmeval}. Math is GSM8K exact match on the final number~\citep{cobbe2021gsm8k}.
Safety uses 100 unsafe and 100 benign XSTest prompts~\citep{rottger2024xstest}; a response counts as a refusal if
its first 300 characters contain one of 27 refusal phrases, and the balanced score averages the refusal rate on
unsafe prompts and the compliance rate on benign ones. Knowledge is MMLU~\citep{hendrycks2021mmlu}, scored by the
likelihood of the answer letters, with expected calibration error~\citep{guo2017calibration} as a secondary metric.

The prompts are zero-shot and model-agnostic, so absolute scores are lower than published few-shot numbers. Llama
and OLMo score only 0.32 and 0.36 on MMLU under this protocol, close to the 0.25 chance level. Every model is
evaluated with the same protocol before and after pruning and healing, so these offsets shift absolute levels but
not comparisons between healing methods.

\begin{table}[t]
  \centering
  \small
  \caption{Evaluation suite. Development items are used only for planning; test items give every reported number.}
  \label{tab:suite}
  \begin{tabular}{llrrl}
    \toprule
    Capability & Benchmark & Test & Dev & Score \\
    \midrule
    Structured output & JSONSchemaBench (GitHub-easy) & 150 & 60 & schema-valid JSON \\
    Function calling & BFCL v3 simple + multiple & 200 & 60 & AST match (name, arguments) \\
    Instruction following & IFEval & 300 & 60 & strict prompt-level accuracy \\
    Mathematical reasoning & GSM8K test & 200 & 60 & exact final answer \\
    Safety refusal & XSTest (100 unsafe, 100 benign) & 200 & 60 & refuse unsafe, answer benign \\
    Knowledge & MMLU (all subjects) & 300 & 60 & answer-letter likelihood; ECE \\
    \bottomrule
  \end{tabular}
\end{table}

\subsection{Healing}

\paragraph{Pools and teacher targets.} Healing data comes from six pools: one per generative capability plus a
general pool. Each pool starts from about 1,300 prompts drawn from sources disjoint from the evaluation sets:
JSON-instance and function-calling prompts built from glaive-function-calling-v2 schemas~\citep{glaive2023functioncalling};
Alpaca-cleaned instructions~\citep{yahma2023alpacacleaned,taori2023alpaca} with one of ten verifiable constraints
attached; the GSM8K training split; harmful PKU-SafeRLHF prompts~\citep{ji2025pkusaferlhf} mixed with benign Alpaca
instructions; and Alpaca-cleaned for the general pool. The unpruned model of the same family writes the targets
greedily (a form of self-data distillation~\citep{thangarasa2025selfdata}), and only outputs that show the skill are
kept: schema-valid JSON, a known function with JSON arguments, a passing IFEval check, or a correct final answer. Up
to 1,000 examples per pool are retained (685--1,000 for math, depending on the family). Every method of a family uses
the same pools and targets. The pools are aligned with the benchmarks, so the recovery we report is recovery on
these evaluations.

\paragraph{Recipe and adaptation scope.} All heals share one recipe: AdamW without weight decay, 5\% linear warm-up
followed by cosine decay, 16 sequences of at most 512 tokens per update, loss on response tokens only, bfloat16
autocast, gradient checkpointing and gradient clipping at 1. A token budget counts every token of the training
sequences, prompt and response, without padding. We compare three adaptation scopes: LoRA~\citep{hu2022lora} of rank
16 ($\alpha = 32$, dropout 0.05) on all attention and MLP projections of every block (\emph{all-LoRA}, learning rate
$2\times10^{-4}$); the same adapters only on the blocks bordering a removed span (\emph{cut-LoRA}); and full
fine-tuning of the last three blocks and the final normalization in 32-bit precision (\emph{last-$k$}, learning rate
$2\times10^{-5}$), adapted from~\citep{lu2026reassessing} but with the output head frozen because it is tied to the
input embeddings in three of the four families. A 1M-token heal takes 239--371 optimizer updates and a 4.48M-token
heal 1,337--1,493.

\subsection{Allocation rules}

A heal is specified by a token vector $\mathbf{t} = (t_p)_{p\in\mathcal{P}}$ over the six pools with
$\sum_p t_p = B$, and a scope. Let $r_c(M) = \mathrm{score}_c(M)/\mathrm{score}_c(M_0)$ be the retention of
capability $c$ relative to the unpruned model $M_0$, and $d_c = 1 - r_c(M_p)$ its damage after pruning. We compare
three rules.

\paragraph{Uniform.} $t_p = B/|\mathcal{P}|$ with all-LoRA.

\paragraph{Damage-proportional.} $t_p \propto \sum_{c:\,\pi(c)=p} \max\{0, d_c\}$, where $\pi$ maps each capability to
its own pool and knowledge to the general pool; damage is measured on the development split; all-LoRA.

\paragraph{RAH.} The planner first runs 21 pilot heals of the pruned model, all scored on the development split:
each pool alone at 30k, 100k and 300k tokens under all-LoRA (18 pilots), and the uniform mixture at 300k tokens under
each of the three scopes (3 pilots). Together they train on 3.48M tokens. The gain in retention of capability $c$
is modeled as a saturating curve of its effective tokens,
\begin{equation}
  \hat r_c(\mathbf{t}) = \min\bigl\{ r_c(M_p) + a_c\bigl(1 - e^{-x_c/\tau_c}\bigr),\; 1.2 \bigr\},
  \qquad x_c = \sum_{p} T_{cp}\, t_p,\quad T_{c\pi(c)} = 1,\; T_{cp} \ge 0,
  \label{eq:curve}
\end{equation}
where $a_c$ is a recovery ceiling, $\tau_c$ a time-scale and $T$ a non-negative transfer matrix. The parameters are
fitted to the pilot gains by bounded, ridge-regularized least squares, with $\tau_c$ no smaller than a tenth of the
smallest pilot budget. RAH then maximizes $\sum_c \hat r_c(\mathbf{t})$ subject to $\sum_p t_p = B$ and a trust
region $t_p \le 600\text{k}$ (twice the largest pilot budget; when $B > 6 \times 600\text{k}$ the region is
infeasible and is relaxed to $B/|\mathcal{P}|$), a concave program solved in milliseconds. The scope
with the highest summed development retention among the three scope pilots is used for the final heal.
\emph{RAH-capped} replaces $\hat r_c$ by $\min\{\hat r_c, 1\}$ in the objective so that predicted retention above
the unpruned model earns nothing. On Qwen it yields the same allocation as RAH (recomputed and recorded in the repository) and was not run
separately. Each
pilot is run once, so the variability of the plan itself is not measured.

\subsection{Token-matched control}

RAH heals with $B = 1$M tokens but also spends 3.48M training tokens on pilots. The token-matched control heals with
the uniform mixture and all-LoRA for 4.48M tokens, the same total. It matches training tokens only. It does not
count the 21 development evaluations of the pilots, which favors RAH, and it does not equalize runtime, since
last-$k$ and LoRA differ in cost per token. At 4.48M tokens every pool is cycled at least 1.5 times.

\subsection{Metrics and statistics}
\label{sec:stats}

Retention can exceed 1 when healing teaches a format the unpruned model handled poorly, and exceeding the unpruned
model is not recovery. The two primary metrics are therefore the \emph{capped mean retention}
$\bar r^{\,\mathrm{cap}} = \frac{1}{|\mathcal{C}|}\sum_c \min\{r_c, 1\}$ and the \emph{worst-case retention}
$r^{\min} = \min_c r_c$. Both are ratios to the unpruned score, not accuracies: a math retention gain of 0.17 on
Llama, whose unpruned GSM8K accuracy is 0.475, is a gain of about 0.08 in accuracy. For the diagnosis we also report
chance-corrected retention $(\mathrm{score}_c - \kappa_c)/(\mathrm{score}_c(M_0) - \kappa_c)$, with $\kappa = 0.25$
for four-option MMLU, $\kappa = 0.5$ for the balanced XSTest score and $\kappa = 0$ otherwise.

Uniform, damage-proportional and RAH are run with three seeds (0, 1, 2) in every family, RAH-capped with three seeds
on Llama, SmolLM2 and OLMo, and the token-matched control with three seeds in every family; other variants use one
seed. We test differences with two paired bootstraps (2,000 resamples). The \emph{item-level} bootstrap averages
each item's score over seeds and resamples items within each capability. The \emph{seed-aware} bootstrap first
resamples each method's seeds and then the items, so it also reflects training randomness~\citep{alahmari2024repeatability};
with three seeds it is coarse but more conservative. Holm's correction is applied within two families of tests,
defined after the results were available: the fixed-budget family (RAH and RAH-capped against uniform and
damage-proportional allocation on both metrics; 28 tests) and the token-matched family (RAH and RAH-capped against
the 4.48M-token uniform heal; 14 tests). A difference is called significant when its Holm-adjusted $p \le 0.05$.
When it is not, we say that no difference was detected; we ran no equivalence tests. A bootstrap $p$ of zero means
$p < 0.001$ (two over the number of resamples), and the Holm values computed from that bound are reported as upper
bounds.

The primary metrics were fixed in a commit (\texttt{73e8582}, dated 5 October 2026) before any RAH test result
existed; at that point the Llama test results of uniform healing (three seeds), damage-proportional allocation (one
seed) and the linear patch were already available. RAH-capped was added after the Llama and SmolLM2 results had been
seen, so OLMo is the only family whose test results did not inform its design. The token-matched control
and the seed-aware analysis were added after the fixed-budget results. We therefore treat the comparisons of RAH with
uniform and damage-proportional allocation as confirmatory and everything else as exploratory.

\subsection{Hardware and software}

The runs used three GPU types. Llama and Qwen were run on RTX 4060 laptop GPUs (8\,GB, PyTorch 2.5.1); Llama's
RAH-capped seeds 1--2 and the 4.48M-token seeds of Llama (seeds 1--2) and Qwen (seeds 0--2) were run on an NVIDIA L4.
SmolLM2 and OLMo were run on an NVIDIA A100 (40\,GB, PyTorch 2.11.0). All runs used transformers
4.56.2~\citep{wolf2020transformers}, PEFT 0.21.2~\citep{mangrulkar2022peft} and lm-evaluation-harness 0.4.13. Within
a family, methods share data, recipe, seeds and test items, but GPU and library differences can shift absolute
scores slightly. The repository records the environment of each family, and its commit history records which runs
were moved to the L4.

\section{Results}

\subsection{What breaks}

\begin{figure}[t]
  \centering
  \includegraphics[width=\linewidth]{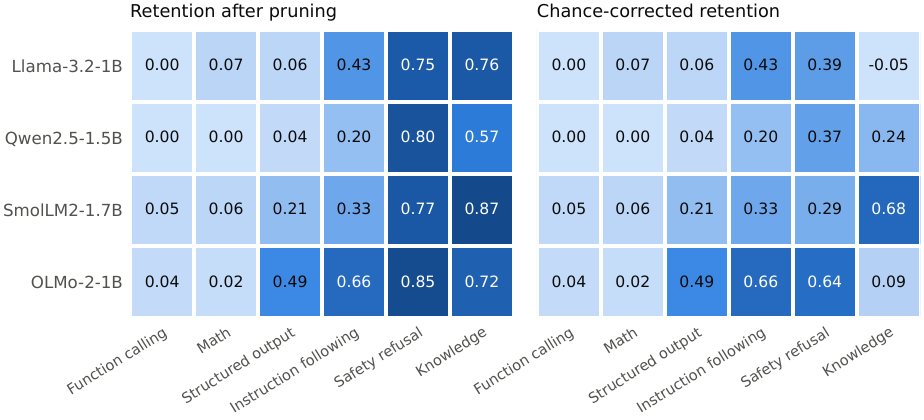}
  \caption{Retention of each capability after the main pruning cut (test split), as raw retention and corrected for
  the chance floors of knowledge and safety. Columns are ordered by mean damage rank. Function calling and math
  collapse in every family.}
  \label{fig:breakage}
\end{figure}

Figure~\ref{fig:breakage} and Table~\ref{tab:damage} show retention after pruning. Function calling (retention
0.00--0.05) and math (0.00--0.07) collapse in every family. For Llama and Qwen most of these failures are structural: after the main
cut, Llama produces parseable JSON for 10\% of structured-output prompts and parseable calls for 6\% of function-calling
prompts, and Qwen for 16\% and 0\%. Structured output and instruction following are damaged in every family but to
very different degrees (0.04--0.49 and 0.20--0.66).

The order in which capabilities break agrees closely across families: Kendall's coefficient of concordance over
the four rankings is $W = 0.92$ (tie-corrected; permutation $p < 0.001$). Safety and knowledge look robust in raw
retention (0.57--0.87), but both have chance floors. After correction, safety retention is 0.29--0.64 and knowledge
ranges from $-0.05$ on Llama, which is at chance, to 0.68 on SmolLM2. With corrected scores the orders still agree,
but only moderately ($W = 0.62$, $p = 0.012$); the mean corrected ranks are 1.62 for function calling, 2.12 for math,
3.25 for structured output, 3.75 for knowledge, 5.00 for safety and 5.25 for instruction following. The robust
conclusion is therefore narrower than the raw one: function calling and math are among the most damaged capabilities
in every family and keep at most 0.07 of their score (on Llama, structured output collapses as far, to 0.06), while
the order of the rest depends on the family. The knowledge result for Llama and OLMo is fragile in any case, because their unpruned MMLU
scores are only 0.07 and 0.11 above chance. A lighter cut shows the same pattern: removing four of Qwen's 28 blocks
already leaves function calling and math at 0.00 and 0.01, whereas Llama keeps at least 0.75 of every capability
except math (0.35) after losing one block.

\begin{table}[t]
  \centering
  \small
  \caption{Damage after the main cut on the test split: unpruned absolute scores, retention (pruned score divided by
  unpruned score) and chance-corrected retention.}
  \label{tab:damage}
  \begin{tabular}{llrrrrrr}
    \toprule
    Family & Quantity & Struct. & Tool & Instr. & Math & Safety & Knowl. \\
    \midrule
    \multirow{3}{*}{Llama-3.2-1B} & unpruned score & 0.32 & 0.54 & 0.45 & 0.47 & 0.83 & 0.32 \\
 & retention & 0.06 & 0.00 & 0.43 & 0.07 & 0.75 & 0.76 \\
 & chance-corrected & 0.06 & 0.00 & 0.43 & 0.07 & 0.39 & -0.05 \\
\midrule
\multirow{3}{*}{Qwen2.5-1.5B} & unpruned score & 0.78 & 0.86 & 0.38 & 0.48 & 0.73 & 0.58 \\
 & retention & 0.04 & 0.00 & 0.20 & 0.00 & 0.80 & 0.57 \\
 & chance-corrected & 0.04 & 0.00 & 0.20 & 0.00 & 0.37 & 0.24 \\
\midrule
\multirow{3}{*}{SmolLM2-1.7B} & unpruned score & 0.71 & 0.69 & 0.45 & 0.39 & 0.74 & 0.43 \\
 & retention & 0.21 & 0.05 & 0.33 & 0.06 & 0.77 & 0.87 \\
 & chance-corrected & 0.21 & 0.05 & 0.33 & 0.06 & 0.29 & 0.68 \\
\midrule
\multirow{3}{*}{OLMo-2-1B} & unpruned score & 0.60 & 0.54 & 0.60 & 0.62 & 0.85 & 0.36 \\
 & retention & 0.49 & 0.04 & 0.66 & 0.02 & 0.85 & 0.72 \\
 & chance-corrected & 0.49 & 0.04 & 0.66 & 0.02 & 0.64 & 0.09 \\
    \bottomrule
  \end{tabular}
\end{table}

\subsection{How you heal matters as much as what you heal with}

\begin{figure}[t]
  \centering
  \includegraphics[width=0.72\linewidth]{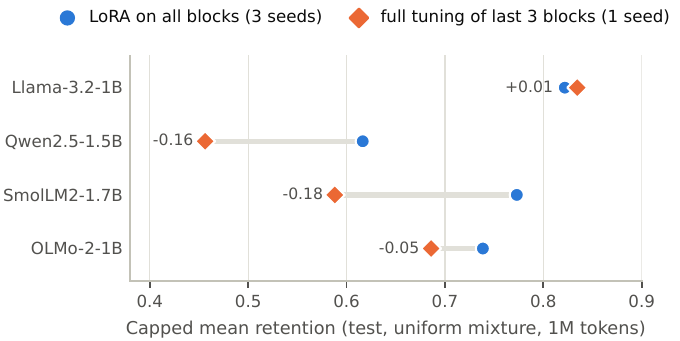}
  \caption{Adaptation scope under identical data. Both heals use the uniform mixture and 1M tokens; only the trained
  parameters differ. Labels give the difference (last three blocks minus all-LoRA).}
  \label{fig:scope}
\end{figure}

Figure~\ref{fig:scope} isolates the adaptation scope: the same uniform mixture and token budget, with either LoRA on
all blocks or full fine-tuning of the last three blocks. On Llama the two are close (0.83 versus 0.82 capped mean),
but on the other three families the last-$k$ scope loses 0.05--0.18, and on Qwen it drops worst-case retention to
0.06. The scope pilots, run on the development split at 300k tokens, rank the scopes the same way in every family:
last-$k$ first on Llama (0.82 versus 0.78 for all-LoRA) and all-LoRA first on Qwen, SmolLM2 and OLMo (0.57, 0.68 and
0.72, against 0.43, 0.47 and 0.64 for last-$k$). The three 300k-token scope pilots together cost less than one 1M-token
heal, and they avoid a loss larger than any difference between allocation rules in this study. The best scope also differs by
capability within a family, so no single scope is best for everything. Our last-$k$ variant keeps the output head
frozen, so this result concerns that adaptation, not the published recipe that also tunes the head.

\subsection{Planning at a fixed budget}

\begin{table}[t]
  \centering
  \footnotesize
  \setlength{\tabcolsep}{3.4pt}
  \caption{Retention on the test split, averaged over three seeds. ``Tokens'' gives the healing tokens and, for RAH,
  the pilot tokens. Bold marks the best capped mean and worst case per family, not significance (Table~\ref{tab:sig}).
  Over-ref.: fraction of benign XSTest prompts refused. On Llama both RAH variants use the last-$k$ scope chosen by
  the scope pilots; all other rows use all-LoRA.}
  \label{tab:main}
  \begin{tabular}{lllrrrrrrrrr}
    \toprule
    Family & Method & Tokens & Struct. & Tool & Instr. & Math & Safety & Knowl. & Capped & Worst & Over-ref. \\
    \midrule
    \multirow{5}{*}{Llama-3.2-1B} & Uniform & 1M & 1.49 & 0.88 & 0.67 & 0.38 & 1.04 & 1.15 & 0.82 & 0.38 & 0.16 \\
 & Damage-prop. & 1M & 2.02 & 0.90 & 0.66 & 0.43 & 0.99 & 1.09 & 0.83 & 0.43 & 0.15 \\
 & RAH & 1M + 3.48M & 1.49 & 0.82 & 0.76 & 0.54 & 0.94 & 1.09 & 0.84 & 0.54 & 0.09 \\
 & RAH-capped & 1M + 3.48M & 1.13 & 0.77 & 0.74 & 0.55 & 0.95 & 1.11 & 0.83 & 0.55 & 0.09 \\
 & Uniform & 4.48M & 1.33 & 0.90 & 0.70 & 0.58 & 1.02 & 1.15 & \textbf{0.86} & \textbf{0.58} & 0.14 \\
\midrule
\multirow{4}{*}{Qwen2.5-1.5B} & Uniform & 1M & 0.73 & 0.54 & 0.58 & 0.21 & 0.96 & 0.67 & 0.62 & 0.21 & 0.59 \\
 & Damage-prop. & 1M & 0.73 & 0.54 & 0.56 & 0.19 & 1.09 & 0.66 & 0.61 & 0.19 & 0.09 \\
 & RAH & 1M + 3.48M & 0.77 & 0.58 & 0.62 & 0.19 & 0.99 & 0.68 & 0.64 & 0.19 & 0.06 \\
 & Uniform & 4.48M & 0.75 & 0.62 & 0.59 & 0.29 & 1.00 & 0.67 & \textbf{0.65} & \textbf{0.29} & 0.52 \\
\midrule
\multirow{5}{*}{SmolLM2-1.7B} & Uniform & 1M & 0.90 & 0.93 & 0.70 & 0.15 & 1.04 & 0.96 & 0.77 & 0.15 & 0.21 \\
 & Damage-prop. & 1M & 0.87 & 0.92 & 0.68 & 0.14 & 1.04 & 0.96 & 0.76 & 0.14 & 0.19 \\
 & RAH & 1M + 3.48M & 0.88 & 0.96 & 0.78 & 0.10 & 1.06 & 0.91 & 0.77 & 0.10 & 0.19 \\
 & RAH-capped & 1M + 3.48M & 0.96 & 0.93 & 0.79 & 0.10 & 0.91 & 0.97 & 0.78 & 0.10 & 0.08 \\
 & Uniform & 4.48M & 0.95 & 0.77 & 0.77 & 0.34 & 0.97 & 0.94 & \textbf{0.79} & \textbf{0.34} & 0.23 \\
\midrule
\multirow{5}{*}{OLMo-2-1B} & Uniform & 1M & 0.77 & 0.97 & 0.71 & 0.21 & 0.96 & 0.81 & 0.74 & 0.21 & 0.16 \\
 & Damage-prop. & 1M & 0.84 & 0.99 & 0.73 & 0.31 & 0.96 & 0.80 & 0.77 & 0.31 & 0.12 \\
 & RAH & 1M + 3.48M & 0.77 & 0.97 & 0.67 & 0.29 & 0.97 & 0.83 & 0.75 & 0.29 & 0.15 \\
 & RAH-capped & 1M + 3.48M & 0.78 & 0.98 & 0.66 & 0.35 & 0.98 & 0.83 & 0.76 & 0.35 & 0.16 \\
 & Uniform & 4.48M & 0.86 & 0.98 & 0.73 & 0.37 & 0.98 & 0.80 & \textbf{0.78} & \textbf{0.37} & 0.16 \\
    \bottomrule
  \end{tabular}
\end{table}

\begin{table}[t]
  \centering
  \small
  \caption{Paired bootstrap tests on the primary metrics. $\Delta$ is the first method minus the second, from per-item
  scores averaged over seeds. $p_{\mathrm{H}}$: Holm-adjusted $p$-value from the item-level bootstrap;
  $p^{\mathrm{seed}}_{\mathrm{H}}$: Holm-adjusted $p$-value from the seed-aware bootstrap; $\le$ marks a bound set by
  the bootstrap resolution.}
  \label{tab:sig}
  \begin{tabular}{llrrrrrr}
    \toprule
    & & \multicolumn{3}{c}{Capped mean} & \multicolumn{3}{c}{Worst case} \\
    \cmidrule(lr){3-5}\cmidrule(lr){6-8}
    Family & Comparison & $\Delta$ & $p_{\mathrm{H}}$ & $p^{\mathrm{seed}}_{\mathrm{H}}$ & $\Delta$ & $p_{\mathrm{H}}$ &
    $p^{\mathrm{seed}}_{\mathrm{H}}$ \\
    \midrule
    \multicolumn{8}{l}{\emph{Fixed budget (1M tokens each; 28 tests)}} \\
Llama-3.2-1B & RAH vs uniform & +0.021 & 1.000 & 1.000 & +0.168 & $\le$0.028 & 0.525 \\
Llama-3.2-1B & RAH vs damage-prop. & +0.013 & 1.000 & 1.000 & +0.116 & 0.460 & 1.000 \\
Llama-3.2-1B & RAH-capped vs uniform & +0.012 & 1.000 & 1.000 & +0.172 & $\le$0.028 & 0.378 \\
Llama-3.2-1B & RAH-capped vs damage-prop. & +0.004 & 1.000 & 1.000 & +0.119 & 0.288 & 1.000 \\
Qwen2.5-1.5B & RAH vs uniform & +0.020 & 1.000 & 1.000 & -0.027 & 1.000 & 1.000 \\
Qwen2.5-1.5B & RAH vs damage-prop. & +0.024 & 0.858 & 1.000 & -0.007 & 1.000 & 1.000 \\
SmolLM2-1.7B & RAH vs uniform & -0.001 & 1.000 & 1.000 & -0.048 & 1.000 & 1.000 \\
SmolLM2-1.7B & RAH vs damage-prop. & +0.008 & 1.000 & 1.000 & -0.043 & 1.000 & 1.000 \\
SmolLM2-1.7B & RAH-capped vs uniform & +0.003 & 1.000 & 1.000 & -0.052 & 1.000 & 1.000 \\
SmolLM2-1.7B & RAH-capped vs damage-prop. & +0.013 & 1.000 & 1.000 & -0.048 & 0.903 & 1.000 \\
OLMo-2-1B & RAH vs uniform & +0.013 & 1.000 & 1.000 & +0.075 & 0.050 & 0.416 \\
OLMo-2-1B & RAH vs damage-prop. & -0.020 & 1.000 & 1.000 & -0.027 & 1.000 & 1.000 \\
OLMo-2-1B & RAH-capped vs uniform & +0.025 & 0.903 & 1.000 & +0.142 & $\le$0.028 & $\le$0.028 \\
OLMo-2-1B & RAH-capped vs damage-prop. & -0.008 & 1.000 & 1.000 & +0.040 & 1.000 & 1.000 \\
\midrule
\multicolumn{8}{l}{\emph{Token-matched (4.48M training tokens each; 14 tests)}} \\
Llama-3.2-1B & RAH vs uniform 4.48M & -0.019 & 1.000 & 1.000 & -0.032 & 1.000 & 1.000 \\
Llama-3.2-1B & RAH-capped vs uniform 4.48M & -0.029 & 0.300 & 0.920 & -0.028 & 1.000 & 1.000 \\
Qwen2.5-1.5B & RAH vs uniform 4.48M & -0.016 & 1.000 & 1.000 & -0.107 & 0.014 & 0.276 \\
SmolLM2-1.7B & RAH vs uniform 4.48M & -0.019 & 1.000 & 1.000 & -0.242 & $\le$0.014 & $\le$0.014 \\
SmolLM2-1.7B & RAH-capped vs uniform 4.48M & -0.014 & 1.000 & 1.000 & -0.247 & $\le$0.014 & $\le$0.014 \\
OLMo-2-1B & RAH vs uniform 4.48M & -0.033 & 0.300 & 0.920 & -0.078 & $\le$0.014 & 0.418 \\
OLMo-2-1B & RAH-capped vs uniform 4.48M & -0.021 & 1.000 & 1.000 & -0.011 & 1.000 & 1.000 \\
    \bottomrule
  \end{tabular}
\end{table}

Table~\ref{tab:main} gives the test results and Table~\ref{tab:sig} the tests. The weakest capability after healing
is math in every run of the methods in Table~\ref{tab:main}, so worst-case retention there is math retention.

\paragraph{Capped mean.} No planned allocation changed capped mean retention significantly on any family. The
largest nominal gain is RAH-capped over uniform healing on OLMo ($+0.025$) and the largest nominal loss is RAH
against damage-proportional allocation on OLMo ($-0.020$). The per-seed standard deviation of capped mean retention is
0.003--0.022, so differences of about 0.02 are within training noise.

\paragraph{Worst case.} RAH and RAH-capped raise math retention over uniform healing on Llama ($+0.17$ for both) and
OLMo ($+0.08$ and $+0.14$); in GSM8K accuracy, Llama rises from 0.178 to 0.260 and OLMo from 0.132 to 0.220 under
RAH-capped. These gains come from allocations that give math about 60\% of the budget, the trust-region maximum
(both variants on Llama, RAH-capped on OLMo). On Llama the gain is confounded with scope, because the planner also chose last-$k$;
the single-seed last-$k$ heal with the uniform mixture already reaches a worst case of 0.49 against 0.38 under
all-LoRA. Under the item-level bootstrap, all four worst-case gains over uniform healing on Llama and OLMo survive
Holm's correction (RAH on OLMo exactly at the threshold, $p_{\mathrm{H}} = 0.050$). Under the seed-aware
bootstrap only one survives: RAH-capped over uniform healing on OLMo ($+0.14$,
$p^{\mathrm{seed}}_{\mathrm{H}} \le 0.028$), the only family whose test results did not inform RAH-capped's design. On Qwen and SmolLM2, where the pilots showed little early math recovery, the planner gave math
only 4--12\% of the budget and its worst case is nominally below uniform healing.

\paragraph{Against damage-proportional allocation.} No difference between the planner and the simpler
damage-proportional rule survives correction. Damage-proportional allocation itself spends nothing on piloting and is
the best 1M-token rule on OLMo by capped mean (0.77).

A simple rule that gives math, the slowest capability, a large share of the budget without any pilots might capture
most of the worst-case gains. We did not test such a rule, so it remains an untested alternative explanation.

\subsection{Planning once the pilots are paid for}

\begin{figure}[t]
  \centering
  \includegraphics[width=\linewidth]{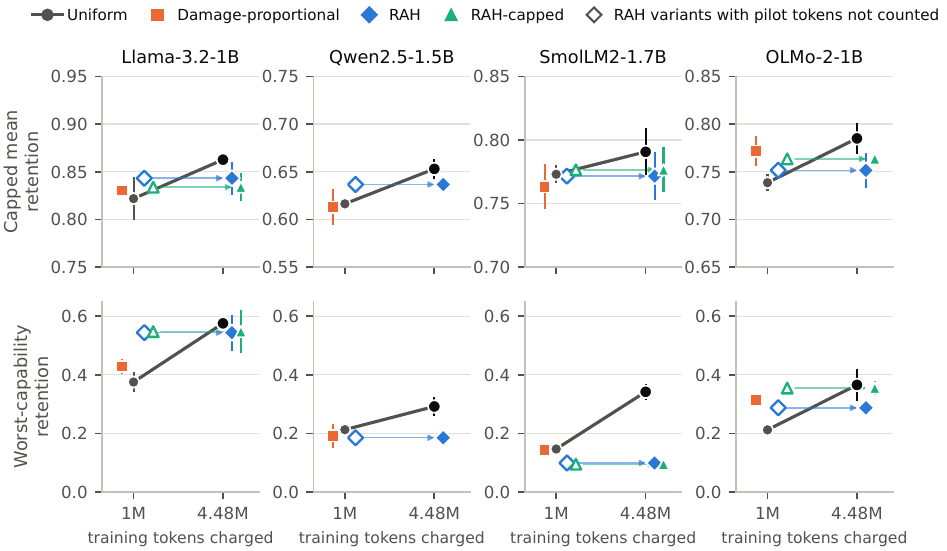}
  \caption{Fixed-budget versus token-matched comparison (three seeds; bars show $\pm1$ seed standard deviation). The
  dark line is uniform healing at 1M and 4.48M tokens. Hollow markers show the RAH variants as they are usually
  reported, at their 1M healing budget; filled markers place them at the 4.48M training tokens they actually used.}
  \label{fig:tm}
\end{figure}

Figure~\ref{fig:tm} moves the planner to the number of training tokens it actually consumed. The token-matched
uniform heal reaches capped means of 0.86, 0.65, 0.79 and 0.78 on Llama, Qwen, SmolLM2 and OLMo, and the best planned
variant reaches 0.84, 0.64, 0.78 and 0.76. Every planned variant has a lower point estimate than the longer uniform
heal on both metrics in every family, though no capped-mean difference is significant after correction. The longer
heal retains math better on Qwen (0.29 versus 0.19), SmolLM2 (0.34 versus 0.10) and OLMo (0.37 versus 0.29 for RAH),
significant after item-level correction; the SmolLM2 differences also survive the seed-aware correction. On Llama and
for OLMo's RAH-capped, no difference was detected.

The extra 3.48M tokens mostly buy math: from 1M to 4.48M tokens of uniform healing, math retention rises from 0.38
to 0.58 on Llama, 0.21 to 0.29 on Qwen, 0.15 to 0.34 on SmolLM2 and 0.21 to 0.37 on OLMo. They are not free of side
effects: on SmolLM2, function-calling retention falls from 0.93 to 0.77, and on Qwen the longer heal keeps the
teacher's over-refusal (52\% of benign prompts refused, against 59\% at 1M tokens). Seed-to-seed standard deviations of the long heal's capped
mean are 0.004--0.018, similar to the 1M-token methods.

A single-seed budget sweep on Qwen at 0.25M and 4M tokens, around the three-seed 1M point, is consistent with this.
In uncapped mean retention RAH is below uniform healing at 0.25M tokens (0.52 versus 0.55) and slightly above at 1M
(0.64 versus 0.62). At 4M the trust region is infeasible and the planned allocation falls back to exactly uniform,
so that point does not test planning (0.66 for both runs).

\subsection{Why planning gains little}

\begin{figure}[t]
  \centering
  \includegraphics[width=\linewidth]{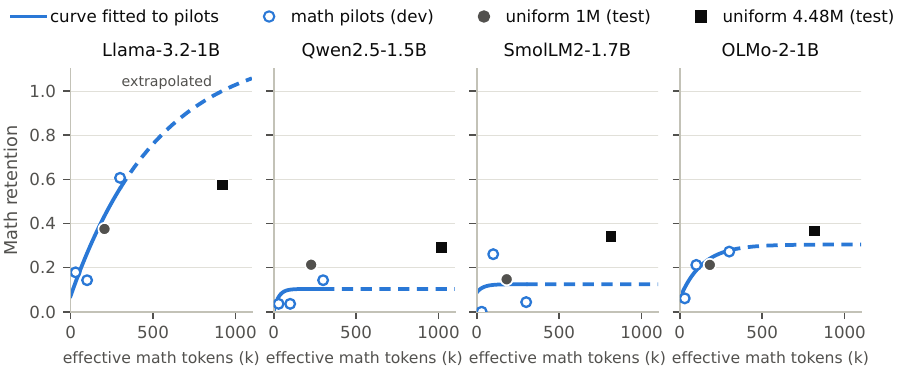}
  \caption{Math recovery curves fitted from the pilots (development split; dashed beyond the largest pilot budget)
  against test retention of uniform heals at 1M and 4.48M tokens, placed at the effective math tokens of
  Equation~\eqref{eq:curve}.}
  \label{fig:curves}
\end{figure}

The planner can only exploit what its pilots see, and the capability that limits the worst case recovers too slowly
for short pilots to measure (Figure~\ref{fig:curves}). On Qwen and SmolLM2 the fitted math curves are fast and
flat, with ceilings of 0.10 and 0.04, yet uniform healing keeps improving math up to 4.48M tokens (0.29 and 0.34).
On Llama the opposite failure occurs: the math curve is slow (time-scale 530k tokens), and its extrapolated ceiling
predicts full math recovery at 4.48M tokens, against an observed 0.58. The pilots' development sets have 60 items
per capability, each pilot is run once, and the transfer matrix cannot represent negative transfer, which some
pilots appear to show, so individual ceilings and time-scales carry substantial uncertainty. In short, cheap pilots cannot tell a slow capability from a
stuck one, and the planner either starves it or over-promises.

\paragraph{Variants and negative results.} The training-free linear patch~\citep{chen2025linearpatch} restores part of
function calling and structured output but trails uniform healing by 0.13--0.28 in capped mean on every family.
Removing the transfer matrix, removing scope selection or switching to a hyperbolic curve changes the capped mean by
at most 0.05 relative to RAH's first seed (one seed each). A max--min objective that targets the worst capability
directly lowers the capped mean below uniform healing on three families. A pilot-free proxy that borrows curve
time-scales from another family produces near-uniform allocations, and its capped mean stays within 0.015 of uniform
healing everywhere.

\subsection{Side effects on refusal}

\begin{figure}[t]
  \centering
  \includegraphics[width=\linewidth]{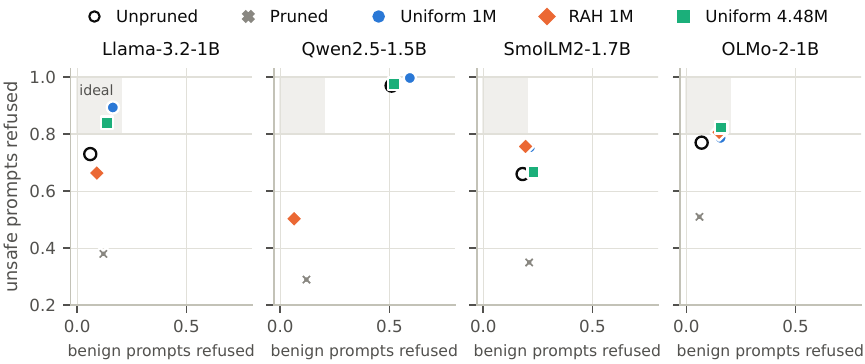}
  \caption{Refusal operating points: fraction of unsafe XSTest prompts refused against fraction of benign prompts
  refused. The ideal is the top-left corner. Pruning lowers refusal of unsafe prompts in every family; healing moves
  the operating point in family-specific directions.}
  \label{fig:refusal}
\end{figure}

Safety has two sides that must be read together: refusing unsafe prompts and complying with benign ones. A heal that
lowers over-refusal is safer only if it does not also refuse fewer unsafe prompts, and in our data, heals that
lowered over-refusal often also refused fewer unsafe prompts. We validated the phrase-matching refusal classifier on
a stratified sample of 240 responses (30 per family and prompt type, drawn from unpruned, pruned and healed models),
labeled by one of the authors from the prompt and response alone, without seeing the model, the method or the
classifier's decision. Against these labels the classifier agrees on 92\% of responses (Cohen's
$\kappa = 0.84$; precision 0.94, recall 0.88), so it slightly under-counts refusals. We also
fixed a bug: the classifier had missed refusals written with a typographic apostrophe, which Llama-3.2 uses. All safety
numbers in this paper are re-scored with the fixed classifier; the planner's Llama inputs were scored before the
fix, so its Llama allocations were planned from the original scores.

Figure~\ref{fig:refusal} shows that pruning lowers the refusal of unsafe prompts in every family (from 0.66--0.97 to
0.29--0.51). Healing moves the operating point in directions that depend on the family and the mixture. On Qwen, the
unpruned model refuses 97\% of unsafe and 51\% of benign prompts; uniform healing, which includes the safety pool,
reproduces this over-refusing teacher at both 1M and 4.48M tokens, whereas RAH assigns no tokens to the safety pool and
refuses 50\% of unsafe and 6\% of benign prompts. On SmolLM2, RAH-capped similarly lowers both rates (to 43\% and
8\%, against 75\% and 21\% under uniform healing). Which operating point is preferable depends on the deployment, and
the balanced score hides the shift, so both rates should always be checked.

\section{Practical Guidance}

For settings like ours (sub-2B instruction-tuned models, Block-Influence cuts and benchmark-aligned healing data):
\begin{itemize}
  \item \textbf{Check function calling and math first.} They collapsed in every family and recovered most slowly, and
  aggregate scores hide their collapse.
  \item \textbf{Run a scope pilot before choosing the mixture.} A wrong scope cost more capped mean retention than
  the 3.48M extra training tokens gained, and LoRA on all blocks was the safer default in three of four families.
  \item \textbf{If the budget is not fixed, heal longer before planning.} A longer uniform heal was competitive with
  every planned variant on every family. Planning is a candidate only when the healing budget itself is fixed, or when
  one fitted plan is reused across many heals so that its pilots are amortized; we did not test such reuse.
  \item \textbf{Report what planning costs.} Compare a planned allocation with a heal that receives the planner's
  pilot tokens, and charge the pilots' evaluations too.
  \item \textbf{Read refusal as two rates.} The share of safety data and the length of the heal both moved the
  balance between refusing unsafe prompts and complying with benign ones.
\end{itemize}

\section{Limitations}

\paragraph{Scope of the study.} All four families have 1.2--1.7B parameters, each is pruned with one criterion at one
main cut, and all data are English. Whether the findings hold for larger models, other pruning criteria or other
languages is untested. We do not measure latency, throughput or memory of the pruned models.

\paragraph{Benchmark-aligned healing.} The pools are close to the benchmarks (for example, the math pool is the GSM8K
training split), so the recovery we report is recovery on these evaluations and may overstate recovery on other tasks
of the same capability. Absolute scores depend on our zero-shot prompts and generation limits.

\paragraph{Statistics.} Test sets of 150--300 items give confidence intervals of about $\pm0.1$ per capability, and
three seeds cannot resolve capped-mean differences much below 0.02. The planner's pilots are run once, so the
variability of its plan is not measured. Several analyses are exploratory, as stated in
Section~\ref{sec:stats}.

\paragraph{Untested baseline.} A pilot-free rule that sends a large share of the budget to the slowest capability
might match the planner's worst-case gains. We did not run it.

\paragraph{Teacher data and hardware.} The teacher targets are regenerated per family, and the recorded statistics
show that SmolLM2's targets were regenerated once between runs (one more instruction-following prompt failed quality
control), so its later runs may have used slightly different targets. Only Llama's teacher files survive in the
repository, together with their checksums. The runs used three GPU types, so absolute scores may differ slightly
between families and between runs moved to another GPU.

\paragraph{Refusal classifier.} The phrase-matching classifier was validated on 240 responses but still misses some
indirect refusals, and we did not judge whether compliant answers to unsafe prompts were actually harmful.

\section{Conclusion}

In four sub-2B instruction-tuned model families, depth pruning reduced function calling and mathematical reasoning
to near zero, while the fate of the other capabilities depended on the family. Healing on benchmark-aligned data
restored much of what broke on these evaluations, and the choice of which parameters to train mattered as much as the
data. A pilot-based planner raised the weakest capability at a fixed budget on two families (robustly on one), but once
its pilot tokens were counted, simply healing longer matched or beat it on every family. Short pilots cannot see how slowly the
hardest capabilities recover. Planned recovery methods should therefore be compared not only at a fixed healing budget
but also against a heal that receives their full training cost. Larger models, other pruning criteria, tasks outside
the healing distribution and a pilot-free math-heavy baseline are the most direct next steps.

\section*{Code and Data Availability}
Code, evaluation items, pool prompts, per-item predictions for every run, the analysis scripts that produce every
table and figure in this paper, and the refusal-validation labels are available at \repourl.

\section*{Acknowledgments}
We thank the developers of the open checkpoints, datasets and software used in this study. AI tools were used for
language editing, code assistance and checking references. The authors designed the study, reviewed all text, code
and results, and take full responsibility for the content.

\bibliographystyle{unsrtnat}
\bibliography{references}

\end{document}